\documentclass{article}

\usepackage[preprint]{neurips_2021}
\usepackage{graphicx}

\usepackage[utf8]{inputenc} 
\usepackage[T1]{fontenc}    
\usepackage{hyperref}       
\usepackage{url}            
\usepackage{booktabs}       
\usepackage{amsfonts}       
\usepackage{nicefrac}       
\usepackage{microtype}      
\usepackage{xcolor}         

\title{Using Contrastive Learning and Pseudolabels to learn representations for Retail Product Image Classification}

\author{
  Muktabh Mayank Srivastava \\
  Paralleldots, Inc. \\
  \texttt{muktabh@paralleldots.com}
  }

\begin{document}

\maketitle

\begin{abstract}
    Retail product Image classification problems are often few shot classification problems, given retail product classes cannot have the type of variations across images like a cat or dog or tree could have. Previous works have shown different methods to finetune Convolutional Neural Networks to achieve better classification accuracy on such datasets. In this work, we try to address the problem statement : Can we pretrain a Convolutional Neural Network backbone which yields good enough representations for retail product images, so that training a simple logistic regression on these representations gives us good classifiers ? We use contrastive learning and pseudolabel based noisy student training to learn representations that get accuracy in order of finetuning the entire Convnet backbone for retail product image classification.
\end{abstract}

\section{Introduction}
Retail product image classification is a computer vision problem frequently encountered in applications like self checkout stores, retail execution measurement, inventory management and manufacturing. A retail product, for example Nutella jar, will hardly have variations among individuals unlike say the category cat, where each individual looks different, so the expectation in most such problems is to be able to train on a minimal number of images. Common real world retail product recognition datasets are often one shot or few shot classification datasets.

\begin{figure}
\centering
  \includegraphics[width=0.9\textwidth]{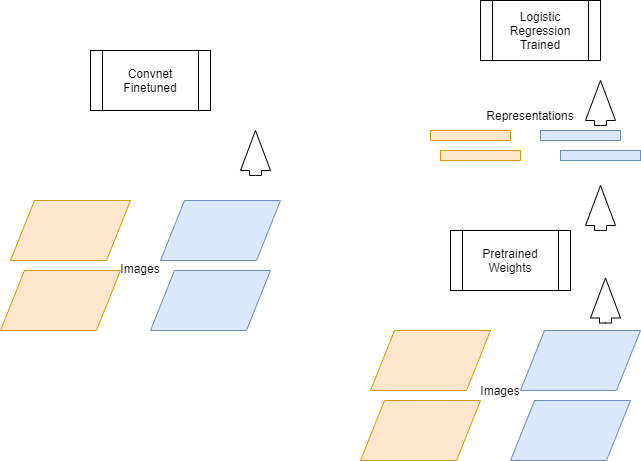}
  \label{fig:training}
  \caption{Previous works need to finetune the entire backbone for training model on a retail image classification dataset. In our work using representations of images from a pretrained model we get equivalent or better accuracy by training just a simple Machine Learning classifier.}
\end{figure}

In our previous work, we had proposed methods to finetune Convolutional Neural Network backbones to classify retail product images. However, given retail products have the property of all individuals of a class looking the same and most of the task of Convnets in such classification problems is to remove real world distortions and noise, one might wonder if a Convnet can be train to create noise invariant image representations that can just be passed through a Logistic Regression or any other simple Machine Learning algorithm to learn recognizing the product. In our work we show that contrastive feature training on a large dataset of image pairs of different retail products [not containing and unrelated to the products we need to train the final classifier on] followed by a noisy pretraining on a large dataset of unannotated retail products, we get a Convolutional backbone whose representations can be passed through a simple Logistic Regression model for classification accuracy almost as good as finetuning a Convnet on images of products we need to classify. Figure ~\ref{fig:training} shows difference between training of between previous works and current method.

\section{Related Work}
In our previous work, we have proposed different tricks to better the accuracy while finetuning Convolutional Neural Networks on Retail Product Image Classification. \citet{Srivastava} We proposed a new layer Local Concepts Accumulation [LCA] layer applied on the output feature map of the Convolutional backbone, which represents an image as a combination of local concepts. There are also published works which finetune GAN-like backbones to recognize Retail Product images using Information Retrieval techniques(\citet{Tonioni2019}). Previously, keypoint matching methods like SIFT (\citet{SIFT} and \citet{6126542}) have also been used to recognize retail products. 

ResNext Convolutional Neural Network backbones (\citet{aggreegated-xie}) pretrained weakly on instagram hashtags and then finetuned on Imagenet (\citet{Mahajan}) have been shown to get better results on Imagenet and on Retail Product images (\citet{Srivastava}).

In more recent times, Contrastive Learning learned representations have shown to perform well for Image classification (\citet{pmlr-v139-zbontar21a}, \citet{chen2020simple}, \citet{khosla2020supervised}, \citet{chen2021exploring}). Even better, these visual representation learners don’t require an annotated dataset and can learn by using an image and its augmentation as training pairs for contrastive learning. However, these algorithms require very large unannotated datasets and need to load a lot of images in GPU memory in a single batch to be able to work. SimSiam which tries to optimize these contrastive learning models to bring down the batchsize can make work at batchsize of 256 as opposed to over 4096 of SIMCLR. 

Noisy student training where a teacher algorithm is used to generate pseudolabels and a student is trained on these pseudolabels has also been used with great results in Computer Vision problems both image classification and object detection (\citet{xie2020selftraining} and \citet{zoph2020rethinking}).

We take the best performing architecture from our experiments in finetuning convnets for retail product image classification which is a ResNext-WSL with a LCA layer and try to create a backbone using it which can be used to learn retail product image representations. Because, it is not possible for us to load large batchsizes of even 256 and train for long periods of time, we use supervised contrative learning with hard example mining on a dataset of annotated image pairs to learn features in the first step as a teacher model. This teacher model is used to produce pseudolabels on a large dataset of unannotated retail product images. In the second step of learning representations, we train a student model as a multitask learning model to learn representations. The two losses in the multitask learning of the student are supervised contrastive loss on an annotated dataset with hard example mining like its teacher and the pseudolabels the teacher algorithm produces on a large unannotated dataset. The representations learnt by both teacher and student are independently analyzed for their performance as input to a Logistic Regression classifier on standard datasets.

\section{Datasets}
\begin{figure}
\centering
  \includegraphics[width=0.4\textwidth]{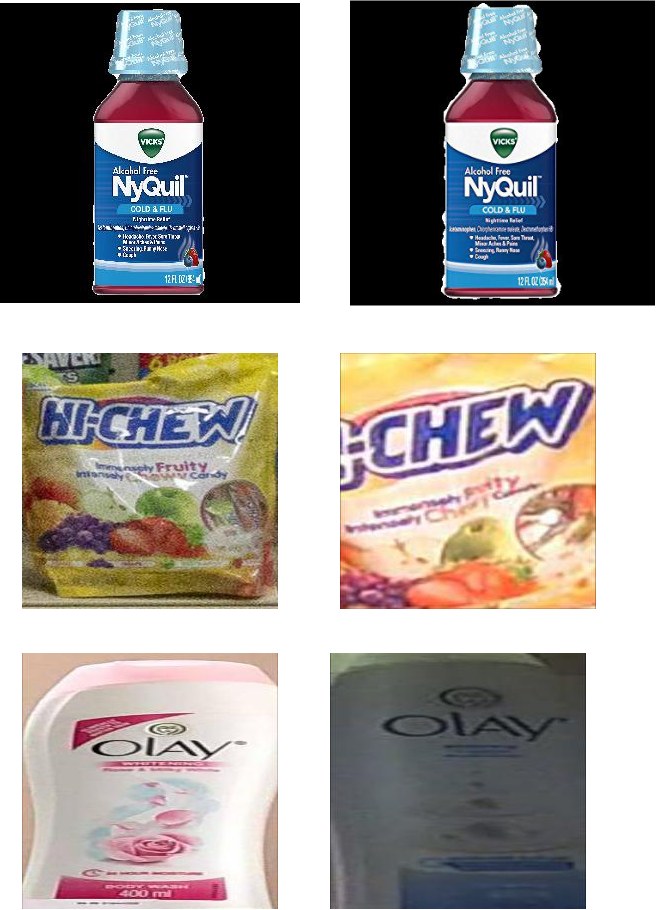}
  \label{fig:pairdatset}
  \caption{Samples from TEACHER-PAIRS dataset. This dataset contains product image pairs crawled from internet and annotated by ParallelDots' annotation team.}
\end{figure}

We first give a description of various datasets used in our work. The first dataset we call TEACHER-PAIRS is an annotated dataset of 250,000 retail product image pairs. This dataset is mined from many other proprietary datasets and crawled from various ecommerce websites. Figure \ref{fig:pairdatset} shows some samples from this dataset.

\begin{figure}
\centering
  \includegraphics[width=0.4\textwidth]{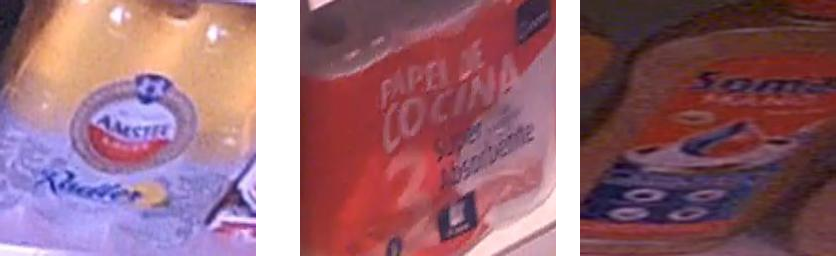}
  \label{fig:pseudodatset}
  \caption{Unannotated samples from STUDENT-PSEUDO dataset.}
\end{figure}

There are no negative annotations, so to train for negative samples, we take random images from outside the pair as a negative sample. A teacher model is trained to learn representations using contrastive loss using hard example mining on the TEACHER-PAIRS dataset. The teacher model is then run over 2 Million unannotated retail product images to generate representations of these images which are treated as pseudolabels. The dataset of unannotated images and their corresponding labels is called STUDENT-PSEUDO.  Student model is then trained on TEACHER-PAIRS with contrastive loss and STUDENT-PSEUDO with Smooth L1 loss. Figure \ref{fig:pseudodatset} shows some samples from dataset.

\begin{figure}
\centering
  \includegraphics[width=0.4\textwidth]{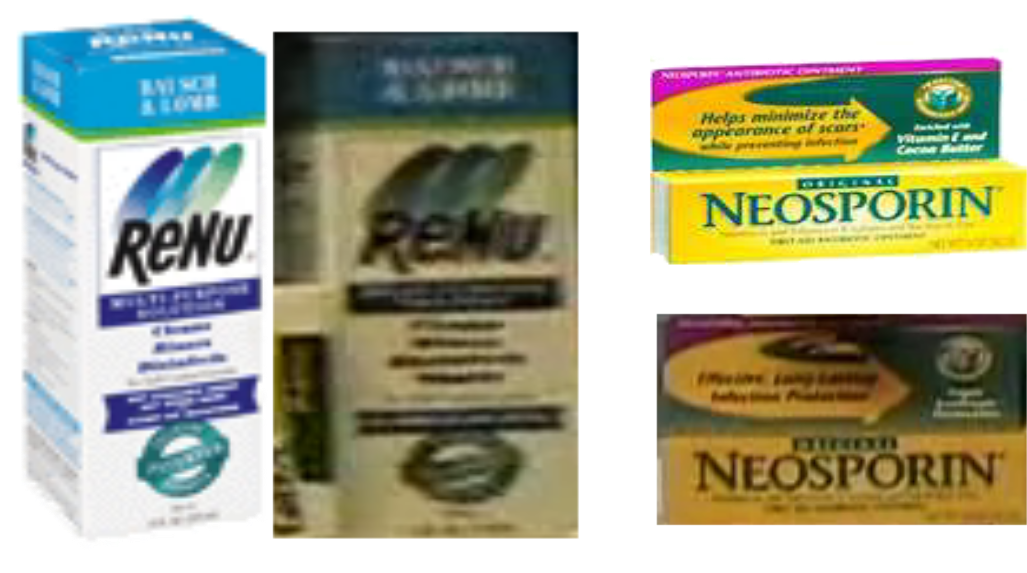}
  \label{fig:grozidatset}
  \caption{Train test sample pair from Grozi-120 dataset.}
\end{figure}

\begin{figure}
\centering
  \includegraphics[width=0.4\textwidth]{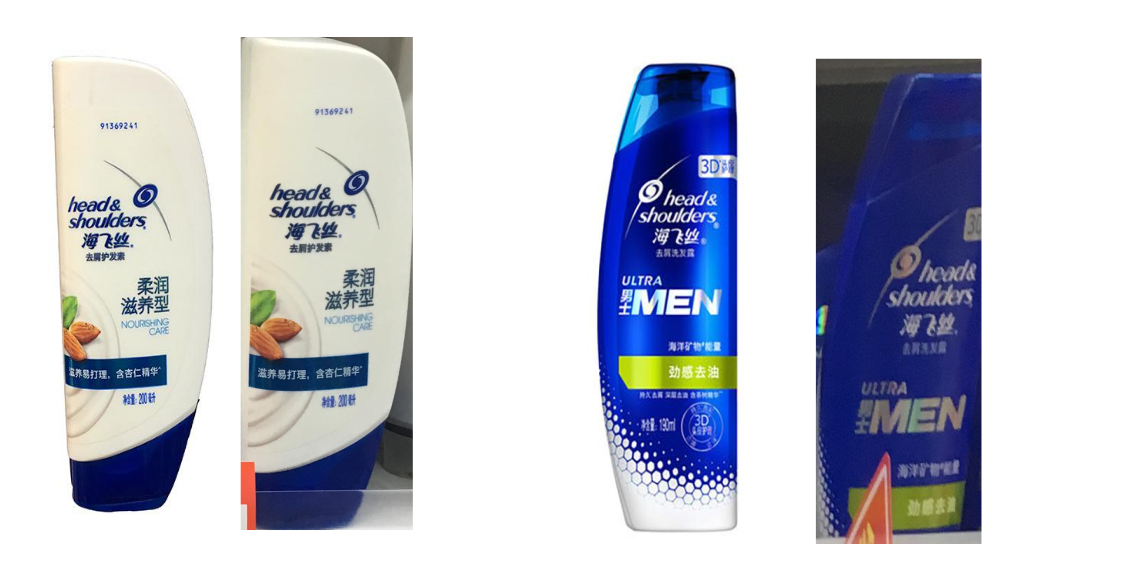}
  \label{fig:capgdatset}
  \caption{Train test sample pair from CAPG-GP dataset.}
\end{figure}

The representations learnt by both teacher and student are tested by creating representations of images in classification subsets of Grozi-120 (\citet{Merler-vitro} and CAPG-GP \citet{Geng-acm}) datasets and training logistic regression classifier on the representations generated. Both Grozi-120 and CAPG-GP are one-shot datasets. Figures \ref{fig:grozidatset} and \ref{fig:capgdatset} show sample train-test pairs from Grozi-120 and CAPG-GP dataset respectively.

\section{Models}
\begin{figure}
\centering
  \includegraphics[width=1.0\textwidth]{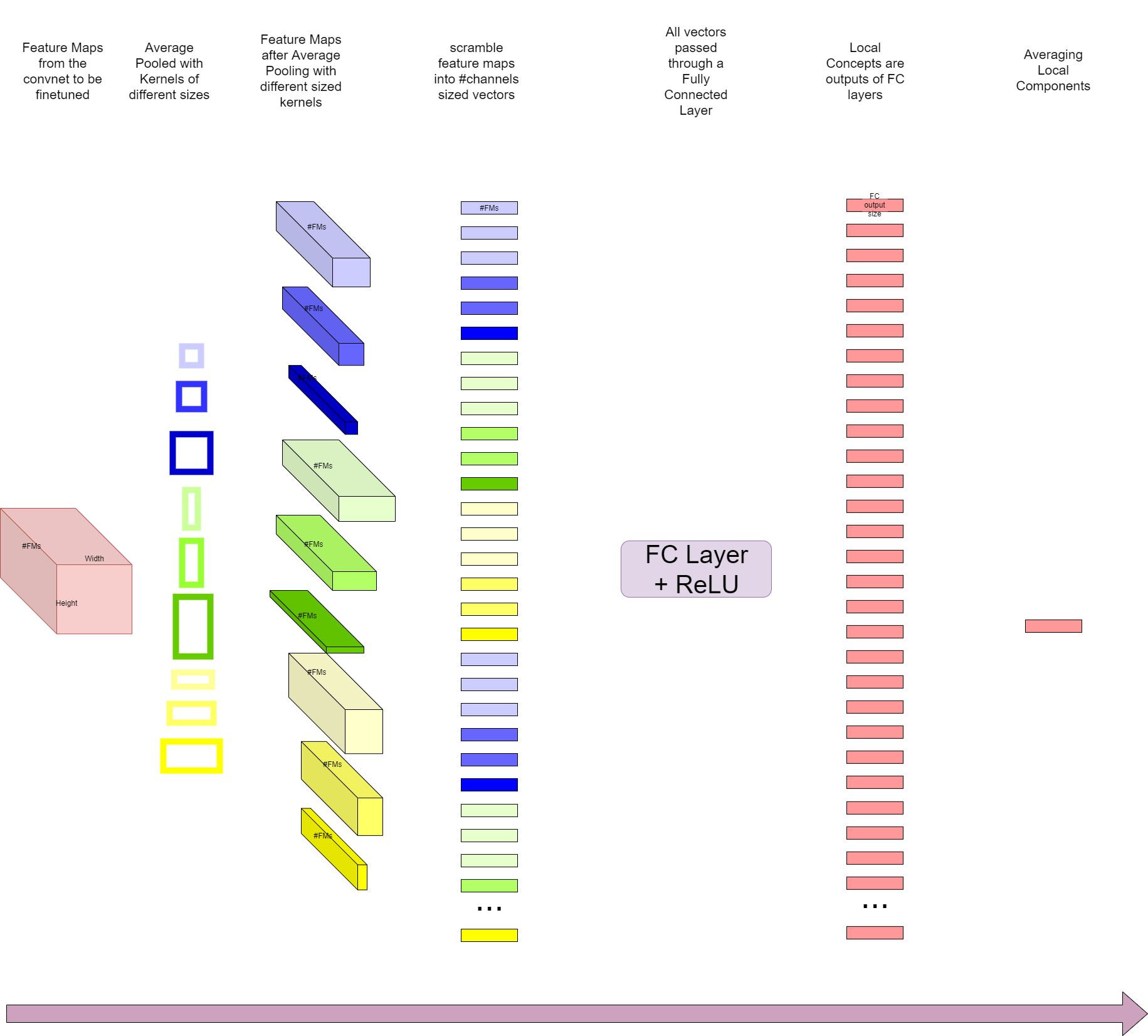}
  \label{fig:lca}
  \caption{LCA layer is places behind a ResNext architecture output feature map to create the representation learning backbone.}
\end{figure}

As noted earlier there are two models we train to learn visual representation. Both have the same Convolutional architecture, which is a Resnext-101\_32X8 architecture. The feature maps of the output of this architecture are passed through a Local Concepts Accumulation layer. Local concepts accumulation (LCA) layer average pools its input feature maps on all rectangular and square sizes larger than 1X1 and creates representations for different local concepts which are then averaged to the representation of the image. The final 2048 dimensional embedding is treated as the representation for the image. LCA layer is same as proposed in our previous work \citet{Srivastava} and is shown in Figure \ref{fig:lca}.

\begin{figure}
\centering
  \includegraphics[width=0.25\textwidth]{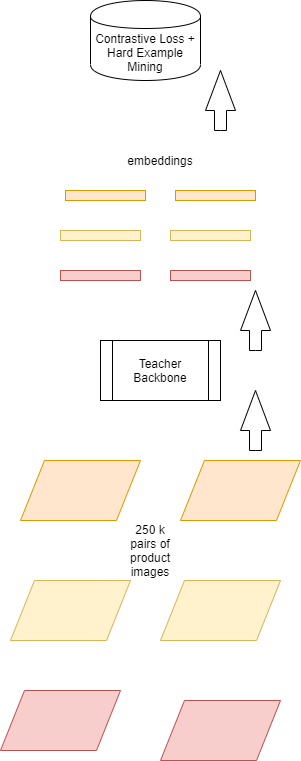}
  \label{fig:teachertraining}
  \caption{Teacher model is trained on the annotated pairs of TEACHER\_PAIRS dataset using contrastive loss and hard example mining. Negative sample for an image is sampled randomly from images outside its pair.}
\end{figure}

In the first step of training, the model is trained on the TEACHER-PAIRS dataset using a contrastive loss function. We use hard example mining to make sure the features learnt are not too simple. The representations this model produces are called Teacher\_Representations. Figure ~\ref{fig:teachertraining} shows training of teacher model.

\begin{figure}
\centering
  \includegraphics[width=1.0\textwidth]{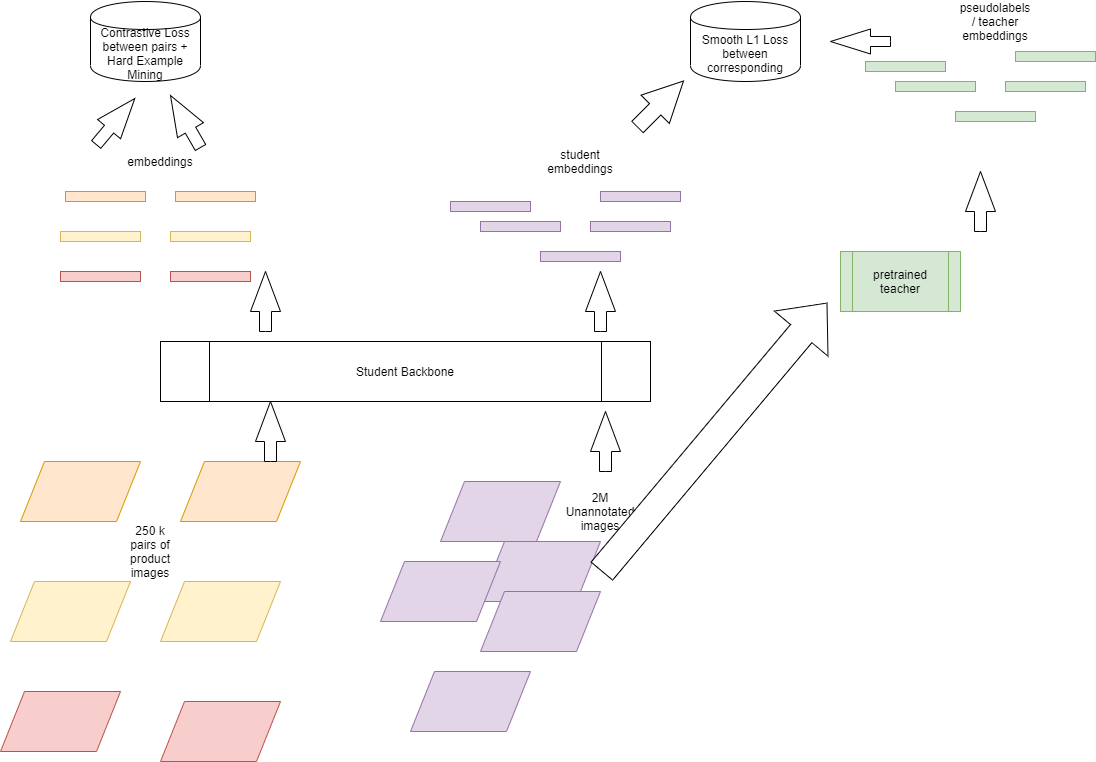}
  \label{fig:studenttraining}
  \caption{Student model is trained on 250 k pairs using contrastive loss and hard example mining and on pseudolabels generated by teacher models on over 2M images. A batch of student model while training half contains supervised image pairs and other half contains unannotated images and their pseudolabels.}
\end{figure}

In the second step of training, the model is trained as a multitask learner on both TEACHER-PAIRS and STUDENT-PSEUDO models. That is, while training, a part of the batch has image pairs from TEACHER-PAIRS and the other part of the batch has images and their representations from STUDENT-PSEUDO. The loss is a weighted average of the contrastive loss on pairs from TEACHER-PAIRS and Smooth L1 loss on STUDENT-PSEUDO representations. The representation from this model is called Student\_Representations. Figure \ref{fig:studenttraining} shows training of student model.

Now for training classifiers for Grozi-120 and CAPG-GP datasets, we first get representations of dataset images from teacher and student models and then pass these representations through a Logistic Regression model to train for classification. We reemphasize that the teacher and student modules are not finetuned, just used to extract representations here.

\section{Results}
We compare the accuracy of the simple Logistic regression model trained on both the teacher and the student representations with our best results of finetuning Convnets for retail product image recognition. 

From our experiments, we conclude that Logistic Regression classifiers trained on the representations derived from the features we learn from TECHER\_PAIRS and STUDENT\_PSEUDO datasets work quite competitively as compared to finetuning entire Convolutional backbone. For Grozi-120 dataset, they work better than finetuning.
\begin{table}\centering
  \caption{Results of various Models on both Datasets}
  \label{sample-table}
  \centering
  \begin{tabular}{p{5cm}rr}
    \toprule
    \cmidrule(l){1-2}
    \textbf{Model Name}     & \textbf{Accuracy [CAPG-GP]}     & \textbf{Accuracy [Grozi-120]}  \\
    \midrule
    \textbf{ResNext-WSL Finetune}  & 84.1\%  & 60.4\%      \\ [0.2cm]
    \textbf{ResNext-WSL with LCA layer Finetune}  & 90.4\%  & 70.8\%   \\ [0.2cm]
    \textbf{ResNext-WSL with LCA layer Finetune with Maximum Entropy Loss}  & 92.2\%  & 72.3\%      \\ [0.2cm]
    \textbf{Teacher\_Represetations + Logistic Regression}  & 87.0\%  & 75.05\%      \\ [0.2cm]
    \textbf{Student\_Represetations + Logistic Regression}  & 87.6\%  & 76.19\%      \\
    \bottomrule
  \end{tabular}
\end{table}

\section{Conclusion}
We show that a visual representation learner which learns on data annotated on any different datasets or crawled from e-commerce websites, modelled as image pairs and combined with unannotated data can be used to learn image representations which can help train very simple and yet accurate classifiers. Retail products keep changing in appearance with new packaging and offers. Finetuning a classifier everytime with addition of new products is costly process. A image representations that allows us to just train logistic regression classifier makes accommodating new product additions very simple.


\newpage


\bibliography{neurips_2021}

\begin{thebibliography}{14}
\expandafter\ifx\csname natexlab\endcsname\relax\def\natexlab#1{#1}\fi

\bibitem[{Chen et~al.(2020)Chen, Kornblith, Norouzi, and
  Hinton}]{chen2020simple}
Ting Chen, Simon Kornblith, Mohammad Norouzi, and Geoffrey Hinton. 2020.
\newblock A simple framework for contrastive learning of visual
  representations.
\newblock \emph{arXiv preprint arXiv:2002.05709}.

\bibitem[{Chen and He(2021)}]{chen2021exploring}
Xinlei Chen and Kaiming He. 2021.
\newblock Exploring simple siamese representation learning.
\newblock In \emph{Proceedings of the IEEE/CVF Conference on Computer Vision
  and Pattern Recognition}, pages 15750--15758.

\bibitem[{Geng et~al.(2018)Geng, Han, Lin, Zhu, Bai, Wang, He, Xiao, and
  Lai}]{Geng-acm}
Weidong Geng, Feilin Han, Jiangke Lin, Liuyi Zhu, Jieming Bai, Suzhen Wang, Lin
  He, Qiang Xiao, and Zhangjiong Lai. 2018.
\newblock \href {https://doi.org/10.1145/3240508.3240522} {Fine-grained grocery
  product recognition by one-shot learning}.
\newblock pages 1706--1714.

\bibitem[{Khosla et~al.(2020)Khosla, Teterwak, Wang, Sarna, Tian, Isola,
  Maschinot, Liu, and Krishnan}]{khosla2020supervised}
Prannay Khosla, Piotr Teterwak, Chen Wang, Aaron Sarna, Yonglong Tian, Phillip
  Isola, Aaron Maschinot, Ce~Liu, and Dilip Krishnan. 2020.
\newblock Supervised contrastive learning.
\newblock \emph{arXiv preprint arXiv:2004.11362}.

\bibitem[{Leutenegger et~al.(2011)Leutenegger, Chli, and Siegwart}]{6126542}
Stefan Leutenegger, Margarita Chli, and Roland~Y. Siegwart. 2011.
\newblock \href {https://doi.org/10.1109/ICCV.2011.6126542} {Brisk: Binary
  robust invariant scalable keypoints}.
\newblock In \emph{2011 International Conference on Computer Vision}, pages
  2548--2555.

\bibitem[{Lowe(2004)}]{SIFT}
D.~G. Lowe. 2004.
\newblock Distinctive image features from scale-invariant keypoints.
\newblock volume~60, pages 91--110.

\bibitem[{Mahajan et~al.(2018)Mahajan, Girshick, Ramanathan, He, Paluri, Li,
  Bharambe, and van~der Maaten}]{Mahajan}
Dhruv Mahajan, Ross Girshick, Vignesh Ramanathan, Kaiming He, Manohar Paluri,
  Yixuan Li, Ashwin Bharambe, and Laurens van~der Maaten. 2018.
\newblock Exploring the limits of weakly supervised pretraining.
\newblock In \emph{Computer Vision -- ECCV 2018}, pages 185--201, Cham.
  Springer International Publishing.

\bibitem[{Merler et~al.(2007)Merler, Galleguillos, and Belongie}]{Merler-vitro}
Michele Merler, Carolina Galleguillos, and Serge Belongie. 2007.
\newblock \href {https://doi.org/10.1109/CVPR.2007.383486} {Recognizing
  groceries in situ using in vitro training data}.
\newblock In \emph{2007 IEEE Conference on Computer Vision and Pattern
  Recognition}, pages 1--8.

\bibitem[{Srivastava(2020)}]{Srivastava}
Muktabh~Mayank Srivastava. 2020.
\newblock Bag of tricks for retail product image classification.
\newblock In \emph{Image Analysis and Recognition}, pages 71--82, Cham.
  Springer International Publishing.

\bibitem[{Tonioni and Stefano(2019)}]{Tonioni2019}
Alessio Tonioni and Luigi~Di Stefano. 2019.
\newblock \href {https://doi.org/10.1016/j.cviu.2019.03.005} {Domain invariant
  hierarchical embedding for grocery products recognition}.
\newblock \emph{Computer Vision and Image Understanding}, 182:81--92.

\bibitem[{Xie et~al.(2020)Xie, Luong, Hovy, and Le}]{xie2020selftraining}
Qizhe Xie, Minh-Thang Luong, Eduard Hovy, and Quoc~V. Le. 2020.
\newblock \href {http://arxiv.org/abs/1911.04252} {Self-training with noisy
  student improves imagenet classification}.

\bibitem[{Xie et~al.(2017)Xie, Girshick, Dollar, Tu, and He}]{aggreegated-xie}
Saining Xie, Ross Girshick, Piotr Dollar, Z.~Tu, and Kaiming He. 2017.
\newblock \href {https://doi.org/10.1109/CVPR.2017.634} {Aggregated residual
  transformations for deep neural networks}.
\newblock pages 5987--5995.

\bibitem[{Zbontar et~al.(2021)Zbontar, Jing, Misra, Lecun, and
  Deny}]{pmlr-v139-zbontar21a}
Jure Zbontar, Li~Jing, Ishan Misra, Yann Lecun, and Stephane Deny. 2021.
\newblock \href {https://proceedings.mlr.press/v139/zbontar21a.html} {Barlow
  twins: Self-supervised learning via redundancy reduction}.
\newblock In \emph{Proceedings of the 38th International Conference on Machine
  Learning}, volume 139 of \emph{Proceedings of Machine Learning Research},
  pages 12310--12320. PMLR.

\bibitem[{Zoph et~al.(2020)Zoph, Ghiasi, Lin, Cui, Liu, Cubuk, and
  Le}]{zoph2020rethinking}
Barret Zoph, Golnaz Ghiasi, Tsung-Yi Lin, Yin Cui, Hanxiao Liu, Ekin~D. Cubuk,
  and Quoc~V. Le. 2020.
\newblock \href {http://arxiv.org/abs/2006.06882} {Rethinking pre-training and
  self-training}.

\end{thebibliography}
\bibliographystyle{neurips_natbib}
\end{document}